# High-Dimensional Covariance Decomposition into Sparse Markov and Independence Domains

**Majid Janzamin** MJANZAMI@UCI.EDU
**Animashree Anandkumar** A.ANANDKUMAR@UCI.EDU
Electrical Engineering and Computer Science, University of California at Irvine, Irvine, CA 92697 USA

## Abstract

In this paper, we present a novel framework incorporating a combination of sparse models in different domains. We posit the observed data as generated from a linear combination of a sparse Gaussian Markov model (with a sparse precision matrix) and a sparse Gaussian independence model (with a sparse covariance matrix). We provide efficient methods for decomposition of the data into two domains, viz., Markov and independence domains. We characterize a set of sufficient conditions for identifiability and model consistency. Our decomposition method is based on a simple modification of the popular $\ell_1$-penalized maximum-likelihood estimator ($\ell_1$-MLE). We establish that our estimator is consistent in both the domains, i.e., it successfully recovers the supports of both Markov and independence models, when the number of samples $n$ scales as $n = \Omega(d^2 \log p)$, where $p$ is the number of variables and $d$ is the maximum node degree in the Markov model. Our conditions for recovery are comparable to those of $\ell_1$-MLE for consistent estimation of a sparse Markov model, and thus, we guarantee successful high-dimensional estimation of a richer class of models under comparable conditions. Our experiments validate these results and also demonstrate that our models have better inference accuracy under simple algorithms such as loopy belief propagation.

**Keywords:** High-dimensional covariance estimation, sparse graphical model selection, sparse covariance models, sparsistency, convex optimization.

## 1. Introduction

Covariance estimation is a classical problem in multivariate statistics. The idea that second-order statistics capture important and relevant relationships between a given set of variables is natural. Finding the sample covariance matrix based on observed data is straightforward and widely used (Anderson, 1984). However, the sample covariance matrix is ill-behaved in high-dimensions, where the number of dimensions $p$ is typically much larger than the number of available samples $n$ ($p \gg n$). Here, the problem of covariance estimation is ill-posed since the number of unknown parameters is larger than the number of available samples, and the sample covariance matrix becomes singular in this regime.

Various solutions have been proposed for high-dimensional covariance estimation. Intuitively, by restricting the class of covariance models to those with a limited number of free parameters, we can successfully estimate the models in high dimensions. A natural mechanism to achieve this is to impose a sparsity constraint on the covariance matrix, which implies that the variables under consideration satisfy *marginal independence*, corresponding to the zero pattern of the covariance matrix (Kauermann, 1996) (and we refer to such models as independence models). In many settings, however, marginal independence is too restrictive and does not hold. For instance, consider the dependence between the monthly stock returns of various companies listed on the S&P 100 index. It is quite possible that a wide range of complex (and unobserved) factors such as the economic climate, interest rates etc., affect the returns of all the companies. Thus, it is not realistic to model the stock returns of various companies through a sparse covariance model.

A popular alternative sparse model, based on *conditional independence* relationships, has gained widespread acceptance in recent years (Lauritzen, 1996). In this case, sparsity is imposed *not* on the covariance matrix, but on the inverse covariance or the *precision* matrix. It can be shown that the zero pattern of the precision matrix corresponds to a set of conditional-independence relationships and such models are referred to as graphical or Markov models. Going back to the stock market example, a first-order approximation is to model the companies in different divisions[1] as conditionally independent given the S&P 100 index variable, which captures the overall trends of the stock returns, and thus removes much of the dependence between the companies in different divisions. However, sparse Markov models may not be always sufficient to capture all the relationships between the variables. Going back to the stock market example, the approximation of using the S&P index node to capture the dependence between companies of different divisions may not be enough. For instance, there can still be a large *residual* dependence between the companies in manufacturing and mining divisions, which cannot be accounted by the S&P index node.

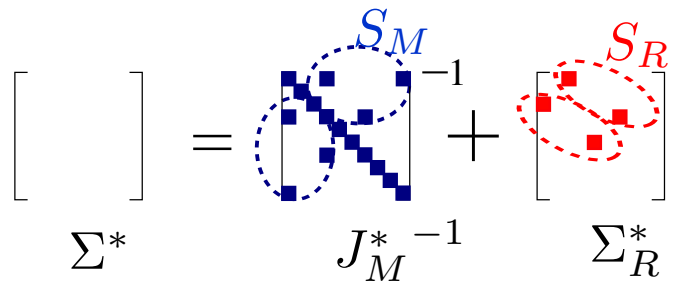

*Figure 1.* Representation of the covariance decomposition problem, where sparsity is imposed on the precision matrix of the first component and the covariance matrix of the second component.

In this paper, we make the above notion precise, and model the variables as a linear combination of samples from an independence and a Markov model. In other words, the covariance matrix of the resulting model is a linear combination of a sparse covariance and a sparse precision matrix, see Fig.1. This forms a richer class of models which can faithfully capture complex relationships, such as in the stock market example above, and yet retain parsimony in the representation.

## Summary of Contributions

We consider joint estimation of Markov and independence models, given observed data in a high dimensional setting. Our contributions in this paper are three fold. First, we derive a set of sufficient restrictions, under which there is a unique decomposition into the two domains, viz., the Markov and the independence domains, thereby leading to an *identifiable* model. Second, we propose novel and efficient estimators for obtaining the decomposition, under both exact and sample statistics. Third, we provide strong theoretical guarantees for high-dimensional learning, both in terms of norm guarantees and *sparsistency* in each domain, viz., the Markov and the independence domain.

Our learning method is based on convex optimization. We adapt the popular $\ell_1$-penalized maximum likelihood estimator (MLE), proposed originally for sparse Markov model selection. This estimator is widely used, and theoretical guarantees on consistent estimation have been proven. Here, an $\ell_1$ penalty is imposed on the precision matrix, which is a convex relaxation of the $\ell_0$ penalty, in order to encourage sparsity in the precision matrix. It is well known that the Lagrangian dual of this program is a *maximum entropy* solution which approximately fits the given sample covariance matrix. We modify this program to our setting as follows: we incorporate an additional $\ell_1$ penalty term involving the residual covariance matrix (corresponding to the independence model) in the max-entropy program. This term can be viewed as encouraging sparsity in the independence domain, while fitting a maximum entropy Markov model to the rest of the sample correlations (after incorporating the independence model). We characterize the optimal solution of the above program, and also provide intuitions on the class of Markov and independence model combinations which can be incorporated under this framework. As a byproduct of this analysis, we obtain a set of conditions for identifiability of the two model components.

We provide strong theoretical guarantees for our proposed method under a set of sufficient conditions. We establish that it is possible to obtain *sparsistency* and norm guarantees in both the Markov and the independence domains, which is somewhat surprising. We establish that the number of samples $n$ is required to scale as $n = \Omega(d^2 \log p)$ for consistency, where $p$ is the number of variables, and $d$ is the maximum degree in the Markov graph. The set of sufficient conditions for successful recovery are based on the so-called notion of *mutual incoherence*, which controls the dependence between different sets of variables, See (Ravikumar et al., 2011). Our conditions are similar to those previously characterized for consistent graphical model selection, and are only slightly stronger. Our consistency proofs borrow ideas from (Ravikumar et al., 2011), and at the same time, require new ideas to carefully control the errors in the two domains, viz., the Markov and the independence domains. This is because the proposed optimization method only ensures that the overall combination of the Markov and the independence models is close to the sample covariance matrix, and does not limit the individual perturbations in the two domains. We consider a careful partitioning of the variables, and impose a set of mutual incoherence conditions, and provide consistency guarantees in high dimensions.

[1] See `http://www.osha.gov/pls/imis/sic_manual.html` for classifications of the companies.

The idea that a combination of Markov and independence models can provide good model-fitting is not by itself new, see (Choi et al., 2010). However, the previous approach has several deficiencies, including lack of theoretical guarantees, assumption of a known sparsity support for the Markov model, use of expectation maximization (EM) which has convergence issues, and so on. In contrast, we develop convex optimization methods for decomposition , and also provide theoretical guarantees for successful recovery. In summary, in this paper, we provide an in-depth study of efficient methods and guarantees for joint estimation of a combination of Markov and independence models.

Our experiments validate our theoretical results and demonstrate that our method is able to learn a richer class of models, compared to sparse graphical model selection, while requiring similar number of samples. In particular, our method is able to provide better estimates for the overall precision matrix, which is dense in general, while the performance of $\ell_1$-based optimization is worse since it attempts to approximate the dense matrix via a sparse estimate. Additionally, we demonstrate that our estimated models have better accuracy under simple distributed inference algorithms such as loopy belief propagation (LBP). This is because the Markov components of the estimated models tend to be more *walk summable* (Malioutov et al., 2006), since some of the correlations can be "transferred" to the residual matrix. Thus, in addition to learning a richer model class, incorporating sparsity in both covariance and precision domains, we also learn models amenable to efficient inference.

## 2. Background and Problem Statement

**Notation:** For any vector $v \in \mathbb{R}^p$ and a real number $a \in [1, \infty)$, the notation $\|v\|_a$ refers to the $\ell_a$ norm of vector $v$ given by $\|v\|_a := \left(\sum_{i=1}^p |v_i|^a\right)^{\frac{1}{a}}$. For any matrix $U \in \mathbb{R}^{p\times p}$, the operator norm is given by $\|\|U\|\|_{a,b} := \max_{\|z\|_a=1} \|Uz\|_b$ for parameters $a, b \in [1, \infty)$. Specifically, we use the $\ell_\infty$ operator norm which is equivalent to $\|\|U\|\|_\infty = \max_{i=1,\dots,p} \sum_{j=1}^p |U_{ij}|$. We also have $\|\|U\|\|_1 = \|\|U^T\|\|_\infty$. We also use the $\ell_\infty$ element-wise norm notation $\|U\|_\infty$ to refer to the maximum absolute value of the entries of matrix $U$. The trace inner product of two matrices is denoted by $\langle U, V\rangle := \mathrm{Tr}(U^T V) = \sum_{i,j} U_{ij}V_{ij}$. Finally, we use the usual notation for asymptotics: $f(n) = \Omega(g(n))$ if $f(n) \geq cg(n)$ for some constant $c > 0$ and $f(n) = O(g(n))$ if $f(n) \leq c'g(n)$ for some constant $c' < \infty$.

### 2.1. Gaussian Graphical Models

A Gaussian graphical model is a family of jointly Gaussian distributions which factor in accordance to a given graph. Given a graph $G = (V, E)$, with $V = \{1, \dots, p\}$, consider a vector of Gaussian random variables $\mathbf{X} = [X_1, X_2, \dots, X_p]$, where each node $i \in V$ is associated with a scalar Gaussian random variable $X_i$. A Gaussian graphical model Markov on $G$ has a probability density function (pdf) that may be parameterized as $f_{\mathbf{X}}(\mathbf{x}) \propto \exp\left[-\frac{1}{2}\mathbf{x}^T J\mathbf{x} + \mathbf{h}^T\mathbf{x}\right]$, where $J$ is a positive-definite symmetric matrix whose sparsity pattern corresponds to that of the graph $G$. More precisely, $J(i, j) = 0$ iff $(i, j) \notin G$. The matrix $J$ is known as the potential or concentration matrix, the non-zero entries $J(i, j)$ as the edge potentials, and the vector $\mathbf{h}$ as the potential vector. This parameterization is known as the information form and is related to the standard mean-covariance parameterization of the Gaussian as $\boldsymbol{\mu} = J^{-1}\mathbf{h}, \quad \Sigma = J^{-1}$, where $\boldsymbol{\mu} := \mathbb{E}[\mathbf{X}]$ is the mean vector and $\Sigma := \mathbb{E}[(\mathbf{X} - \boldsymbol{\mu})(\mathbf{X} - \boldsymbol{\mu})^T]$ is the covariance matrix.

We say that a jointly Gaussian random vector $\mathbf{X}$ with joint pdf $f(\mathbf{x})$ satisfies local Markov property with respect to a graph $G$ if $f(x_i|\mathbf{x}_{\mathcal{N}(i)}) = f(x_i|\mathbf{x}_{V\setminus i})$ holds for all nodes $i \in V$, where $\mathcal{N}(i)$ denotes the set of neighbors of node $i \in V$ and, $V \setminus i$ denotes the set of all nodes excluding $i$. More generally, we say that $\mathbf{X}$ satisfies the global Markov property, if for all disjoint sets $A, B \subset V$, we have

$$f(\mathbf{x}_A, \mathbf{x}_B|\mathbf{x}_S) = f(\mathbf{x}_A|\mathbf{x}_S)f(\mathbf{x}_B|\mathbf{x}_S). \tag{1}$$

where set $S$ is a *separator*[2] of $A$ and $B$. The local and global Markov properties are equivalent for non-degenerate Gaussian distributions (Lauritzen, 1996).

In this paper, we consider models which are characterized by a combination of Markov and independence graphs. In particular, we model the covariance matrix as a linear combination of Markov and independence models:

$$\Sigma^* = {J_M^*}^{-1} + \Sigma_R^*, \quad \mathrm{Supp}(J_M^*) = G_M, \mathrm{Supp}(\Sigma_R^*) = G_R, \tag{2}$$

where $\mathrm{Supp}(\cdot)$ denotes the set of non-zero (off-diagonal) entries, $G_M$ denotes the Markov graph and $G_R$, the independence graph.

[2]A set $S \subset V$ is a separator for sets $A$ and $B$ if the removal of nodes in $S$ partitions $A$ and $B$ into distinct components.

### 2.2. Problem Statement

We now give a detailed description of our problem statement, which consists of the covariance decomposition problem (given exact statistics) and covariance estimation problem (given a set of samples).

**Covariance Decomposition Problem:** A fundamental question to be addressed is the identifiability of the model parameters.

**Definition 1** (Identifiability). *A parametric model $\{P_\theta : \theta \in \Theta\}$ is identifiable with respect to a measure $\mu$ if there do not exist two distinct parameters $\theta_1 \neq \theta_2$ such that $P_{\theta_1} = P_{\theta_2}$ almost everywhere w.r.t. to $\mu$.*

Thus, if a model is not identifiable, there is no hope of estimating the model parameters from observed data. A Gaussian graphical model (with no hidden variables) belongs to the family of standard exponential distributions (Wainwright & Jordan, 2008). Under non-degeneracy conditions, it is also in the minimal form, and as such is identifiable (Brown, 1986). In our setting in (2), however, identifiability is not straightforward to address, and forms an important part of the covariance decomposition problem, described below.

**Decomposition Problem:** Given the covariance matrix $\Sigma^* = {J_M^*}^{-1} + \Sigma_R^*$ as in (2), where $J_M^*$ is an unknown concentration matrix and $\Sigma_R^*$ is an unknown residual covariance matrix, how and under what conditions can we uniquely recover $J_M^*$ and $\Sigma_R^*$ from $\Sigma^*$?

In other words, we want to address whether the matrices $J_M^*$ and $\Sigma_R^*$ are *identifiable*, given $\Sigma^*$, and if so, how can we design efficient methods to recover them. If we do not impose any additional restrictions, there exists an *equivalence class* of models which form solutions to the decomposition problem. For instance, we can model $\Sigma^*$ entirely through an independence model ($\Sigma^* = \Sigma_R^*$), or through a Markov model ($\Sigma^* = {J_M^*}^{-1}$). However, in most scenarios, these extreme cases are not desirable, since they result in dense models, while we are interested in sparse representations with a parsimonious use of edges in both the graphs, viz., the Markov and the independence graphs. In Section 3.1, we provide a sufficient set of structural and parametric conditions to guarantee identifiability of the Markov and the independence components, and in Section 3.2, we propose an optimization program to obtain them.

**Covariance Estimation Problem:** In the above decomposition problem, we assume that the exact covariance matrix $\Sigma^*$ is known. However, in practice, we only have access to samples, and we describe this setting below.

Denote $\widehat{\Sigma}^n$ as the sample covariance matrix $\widehat{\Sigma}^n := \frac{1}{n}\sum_{k=1}^n x_{(k)} x_{(k)}^T$, where $x_{(k)}, k = 1, ..., n$ are $n$ i.i.d. observations of a zero mean Gaussian random vector $X \sim \mathcal{N}(0, \Sigma^*)$, where $X := (X_1, ..., X_p)$. Now the estimation problem is described below.

**Estimation Problem:** Assume that there exists a unique decomposition $\Sigma^* = {J_M^*}^{-1} + \Sigma_R^*$ where $J_M^*$ is an unknown concentration matrix with bounded entries and $\Sigma_R^*$ is an unknown sparse residual covariance matrix given a set of constraints. Given the sample covariance matrix $\widehat{\Sigma}^n$, our goal is to find estimates of $J_M^*$ and $\Sigma_R^*$ with provable guarantees.

In the sequel, we relate the exact and the sample versions of the decomposition problem. In Section 4, we propose a modified optimization program to obtain efficient estimates of the Markov and independence components. Under a set of sufficient conditions, we provide guarantees in terms of *sparsistency*, *sign consistency*, and *norm* guarantees, defined below.

**Definition 2** (Estimation Guarantees). *We say that an estimate $(\widehat{J}_M, \widehat{\Sigma}_R)$ to the decomposition problem in (2), given a sample covariance matrix $\widehat{\Sigma}^n$, is sparsistent or model consistent, if the supports of $\widehat{J}_M$ and $\widehat{\Sigma}_R$ coincide with the supports of $J_M^*$ and $\Sigma_R^*$ respectively. It is said to be sign consistent, if additionally, the respective signs coincide. The norm guarantees on the estimates is in terms of bounds on $\|\widehat{J}_M - J_M^*\|$ and $\|\widehat{\Sigma}_R - \Sigma_R^*\|$, under some norm $\|\cdot\|$.*

## 3. Analysis under Exact Statistics

### 3.1. Assumptions under Exact Statistics

We first provide a set of sufficient conditions under which we can guarantee that the decomposition of $\Sigma^*$ in (2) into concentration matrix $J_M^*$ and residual matrix $\Sigma_R^*$ is unique. We impose the following set of constraints on the two matrices:

(A.0) $J_M^*$ is a positive definite matrix: $J_M^* \succ 0$.

(A.1) Off-diagonal entries of $J_M^*$ are bounded from above, i.e., $\|J_M^*\|_{\infty,\text{off}} \leq \lambda^*$, for some $\lambda^* > 0$.

(A.2) Diagonal entries of $\Sigma_R^*$ are zero: $(\Sigma_R^*)_{ii} = 0$, and the support of its off-diagonal entries satisfies

$$\left(\Sigma_R^*\right)_{ij} \neq 0 \iff |\left(J_M^*\right)_{ij}| = \lambda^*, \quad \forall\, i \neq j. \quad (3)$$

(A.3) For any $i, j$, we have $\text{sign}\big(\left(\Sigma_R^*\right)_{ij}\big).\,\text{sign}\big(\left(J_M^*\right)_{ij}\big) \leq 0$, i.e., the signs are opposite to one another.

In the sequel, we propose an efficient method to recover the respective matrices $J_M^*$ and $\Sigma_R^*$ under conditions (A.0)-(A.3) and then establish the uniqueness of the decomposition. Finally, note that we do not impose any sparsity constraints on the concentration matrix $J_M^*$, and in fact, our method and guarantees allow for dense matrices $J_M^*$, when the exact covariance matrix $\Sigma^*$ is available. However, when only samples are available, we limit ourselves to sparse $J_M^*$ and provide learning guarantees in the high-dimensional regime, where the number of samples can be much smaller than the number of variables.

### 3.2. Formulation of the Optimization Program

We now propose a method based on convex optimization for obtaining $(J_M^*, \Sigma_R^*)$ given the covariance matrix $\Sigma^*$ in (2). Consider the following program

$$\big(\widehat{\Sigma}_M, \widehat{\Sigma}_R\big) := \arg\max_{\Sigma_M \succ 0, \Sigma_R} \log\det \Sigma_M - \lambda \|\Sigma_R\|_{1,\text{off}}$$
$$\text{s.t.}\quad \Sigma_M + \Sigma_R = \Sigma^*,\ (\Sigma_R)_d = 0, \qquad (4)$$

where $\|\cdot\|_{1,\text{off}}$ denotes the $\ell_1$ norm of the off-diagonal entries, which is the sum of the absolute values of the off-diagonal entries, and $(\cdot)_d$ denotes the diagonal entries. Intuitively, the parameter $\lambda$ imposes a penalty on large residual covariances, and under favorable conditions, can encourage sparsity in the residual matrix. The program in (4) can be recast

$$\big(\widehat{\Sigma}_M, \widehat{\Sigma}_R\big) := \arg\max_{\Sigma_M \succ 0, \Sigma_R} \log\det \Sigma_M \qquad (5)$$
$$\text{s.t.}\quad \Sigma_M + \Sigma_R = \Sigma^*,\ (\Sigma_R)_d = 0, \|\Sigma_R\|_{1,\text{off}} \le C(\lambda),$$

for some constant $C(\lambda)$ depending on $\lambda$. The objective function in the above program corresponds to the entropy of the Markov model (modulo a scaling and a shift factor) (Cover & Thomas, 2006), and thus, intuitively, the above program looks for the optimal Markov model with maximum entropy subject to an $\ell_1$ constraint on the residual matrix.

We declare the optimal solution $\widehat{\Sigma}_R$ in (4) as the estimate of the residual matrix $\Sigma_R^*$, and $\widehat{J}_M := \widehat{\Sigma}_M^{-1}$ as the estimate of the Markov concentration matrix $J_M^*$. The justification behind these estimates is based on the fact that the Lagrangian dual of the program in (4) is (see long version on arXiv)

$$\widehat{J}_M := \arg\min_{J_M \succ 0} \langle \Sigma^*, J_M\rangle - \log\det J_M \qquad (6)$$
$$\text{s.t.}\ \|J_M\|_{\infty,\text{off}} \le \lambda,$$

where $\|\cdot\|_{\infty,\text{off}}$ denotes the $\ell_\infty$ element-wise norm of the off-diagonal entries, which is the maximum absolute value of the off-diagonal entries. Further, we show in the long version that the following relations exist between the optimal primal [3] solution $\widehat{J}_M$ and the optimal dual solution $\big(\widehat{\Sigma}_M, \widehat{\Sigma}_R\big)$: $\widehat{J}_M = \widehat{\Sigma}_M^{-1}$, and thus $\widehat{J}_M^{-1} + \widehat{\Sigma}_R = \Sigma^*$ is a valid decomposition of the covariance matrix $\Sigma^*$.

[3] Henceforth, we refer to the program in (6) as the primal program and the program in (4) as the dual program.

**Remark:** Notice that when the $\ell_\infty$ constraint is removed in the primal program in (6), which is equivalent to letting $\lambda \to \infty$, the program corresponds to the maximum likelihood estimate, and the optimal solution in this case is $\widehat{J}_M = {\Sigma^*}^{-1}$ and $\widehat{\Sigma}_R = 0$. At the other extreme, when $\lambda \to 0$, $\widehat{J}_M$ is a diagonal matrix, and the residual matrix $\widehat{\Sigma}_R$ is in general, a full matrix. (except for the diagonal entries). Thus, the parameter $\lambda$ allows us to carefully tune the contributions of the Markov and residual components.

### 3.3. Guarantees and main results

The main decomposition result is as follows. The proofs can be found in the extended version on arXiv.

**Theorem 1** (Uniqueness of Decomposition)**.** *Under (A.0)–(A.3), given a covariance matrix $\Sigma^*$, if we set the parameter $\lambda = \|J_M^*\|_{\infty,\text{off}}$ in the optimization program in* (4)*, then the optimal solutions of primal-dual optimization programs* (6) *and* (4) *are given by $\big(\widehat{J}_M, \widehat{\Sigma}_R\big) = \big(J_M^*, \Sigma_R^*\big)$, and the decomposition is unique.*

Thus, we establish that the proposed optimization programs in (4) and (6) *uniquely* recover the Markov concentration matrix $J_M^*$ and the residual covariance matrix $\Sigma_R^*$ given $\Sigma^*$ under conditions (A.0)–(A.3).

## 4. Sample Analysis of the Algorithm

### 4.1. Optimization Program

We have so far provided guarantees on unique decomposition given the exact covariance matrix $\Sigma^*$. We now consider the case, when $n$ i.i.d. samples are available from $\mathcal{N}(0, \Sigma^*)$. We now modify the primal-dual pair (6) and (4), considered in the previous section, to incorporate the sample covariance matrix $\widehat{\Sigma}^n$.

$$\widehat{J}_M := \arg\min_{J_M \succ 0} \langle \widehat{\Sigma}^n, J_M\rangle - \log\det J_M + \gamma \|J_M\|_{1,\text{off}}$$
$$\text{s.t.}\ \|J_M\|_{\infty,\text{off}} \le \lambda, \qquad (7)$$

It is shown that the dual of above program is

$$\big(\widehat{\Sigma}_M, \widehat{\Sigma}_R\big) := \arg\max_{\Sigma_M \succ 0, \Sigma_R} \log\det \Sigma_M - \lambda \|\Sigma_R\|_{1,\text{off}}$$
$$\text{s.t.}\ \|\widehat{\Sigma}^n - \Sigma_M - \Sigma_R\|_{\infty,\text{off}} \le \gamma, \qquad (8)$$
$$\big(\Sigma_M\big)_d = \big(\widehat{\Sigma}^n\big)_d,\ \big(\Sigma_R\big)_d = 0.$$

We further establish that $\widehat{\Sigma}_M = \widehat{J}_M^{-1}$ and thus,

$$\|\widehat{\Sigma}^n - \widehat{J}_M^{-1} - \widehat{\Sigma}_R\|_{\infty,\text{off}} \le \gamma. \qquad (9)$$

Comparing the above with the exact decomposition $\Sigma^* = {J_M^*}^{-1} + \Sigma_R^*$ in (2), we note that for the sample version, we do not exactly fit the Markov and the residual models with the sample covariance matrix $\widehat{\Sigma}^n$, but allow for some divergence, depending on $\gamma$. Similarly, the primal program in (7) has an additional $\ell_1$ penalty term on $\widehat{J}_M$, which is absent in (6). Having a non-zero $\gamma$ in the above programs enables us to impose a sparsity constraint on $\widehat{J}_M$, which in turn, enables us to estimate the matrices in the high dimensional regime, under a set of sufficient conditions given below.

### 4.2. Assumptions under Sample Statistics

The additional assumptions for successful recovery in high dimensions are based on the Hessian of the objective function in the optimization program in (7), with respect to the variable $J_M$, evaluated at the true Markov model $J_M^*$. The Hessian of this function is given by (Boyd & Vandenberghe, 2004)

$$\Gamma^* = {J_M^*}^{-1} \otimes {J_M^*}^{-1} = \Sigma_M^* \otimes \Sigma_M^*, \tag{10}$$

where $\otimes$ denotes the Kronecker matrix product. Thus $\Gamma^*$ is a $p^2 \times p^2$ matrix indexed by the node pairs. Based on the results for exponential families (Brown, 1986), $\Gamma^*_{(i,j),(k,l)} = \mathrm{Cov}\{X_iX_j, X_kX_l\}$, and hence it can be interpreted as an edge-based alternative to the usual covariance matrix $\Sigma_M^*$. Define $K_M$ as the $\ell_\infty$ operator norm of the covariance matrix of the Markov model: $K_M := \|\!|\Sigma_M^*|\!\|_\infty$. We now denote the supports of the Markov and residual models. Denote $E_M := \{(i,j) \in V \times V | i \neq j, \left(J_M^*\right)_{ij} \neq 0\}$ as the edge set of Markov matrix $J_M^*$. Define

$$S_M := E_M \cup \{(i,i) | i = 1, ..., p\}, \tag{11}$$

$$S_R := \{(i,j) \in V \times V | \left(\Sigma_R^*\right)_{ij} \neq 0\}. \tag{12}$$

Thus, the set $S_M$ includes diagonal entries and also all edges of the Markov graph corresponding to $J_M^*$. Also, recall from (A.2) that the diagonal entries of $\Sigma_R^*$ are set to zero, and that the support set $S_R$ is contained in $S_M$, i.e., $S_R \subset S_M$. Let $S_M^c$ and $S_R^c$ denote the respective complement sets. Define $S := S_M \cap S_R^c$, so that $\{S_R, S, S_M^c\}$ forms a partition of $\{(1, ..., p) \times (1, ..., p)\}$. This partitioning plays a crucial role in being able to provide learning guarantees. Define the maximum node degree for Markov model $J_M^*$ as

$$d := \max_{j=1,\dots,p} |\{i : (i,j) \in E_M\}|. \tag{13}$$

Finally, for any two subsets $T$ and $T'$ of $V \times V$, $\Gamma^*_{TT'}$ denotes the submatrix of $\Gamma^*$ indexed by $T$ as rows and $T'$ as columns. We now impose various constraints on the submatrices of the Hessian in (10), limited to each of the sets $\{S_R, S, S_M^c\}$.

(A.4) **Mutual Incoherence**: These conditions impose mutual incoherence among three partitions of $\Gamma^*$ indexed by $S_R$, $S_M^c$ and $S$.

$$\begin{aligned}\max\{&\|\!|\Gamma^*_{S_M^c S}\left(\Gamma^*_{SS}\right)^{-1}\Gamma^*_{SS_R} - \Gamma^*_{S_M^c S_R}|\!\|_\infty,\\ &\|\!|\Gamma^*_{S_M^c S}\left(\Gamma^*_{SS}\right)^{-1}|\!\|_\infty\} \leq (1-\alpha)\\ &\text{for some } \alpha \in (0,1],\end{aligned} \tag{14}$$

$$K_{SS_R} := \|\!|\left(\Gamma^*_{SS}\right)^{-1}\Gamma^*_{SS_R}|\!\|_\infty < \frac{1}{4}. \tag{15}$$

(A.5) **Covariance control**: For the same $\alpha$ specified above, we have the bound:

$$K_{SS} := \|\!|\left(\Gamma^*_{SS}\right)^{-1}|\!\|_\infty \leq \frac{(m-4)\alpha}{4(m-(m-1)\alpha)} \quad \text{for some } m > 4. \tag{16}$$

Assumption (A.4) controls the pairwise effects of edges in different sets $S$, $S_R$ and $S_M^c$ to each other.

### 4.3. Guarantees and Main Results

We are now ready to provide the main result.

**Theorem 2.** *Consider a Gaussian distribution with covariance matrix $\Sigma^* = {J_M^*}^{-1} + \Sigma_R^*$ satisfying conditions (A.0)-(A.5). Given a sample covariance matrix $\widehat{\Sigma}^n$ using $n$ i.i.d. samples from the Gaussian model, let $\left(\widehat{J}_M, \widehat{\Sigma}_R\right)$ denote the unique optimal solutions of the primal-dual pair* (7) *and* (8)*, with parameters $\gamma = C_1\sqrt{\frac{\log p}{n}}$ and $\lambda = \lambda^* + C_2\sqrt{\frac{\log p}{n}}$ for some constants $C_1, C_2 > 0$, where $\lambda^* := \|J_M^*\|_{\infty,\mathrm{off}}$. Suppose that $\left(\Sigma_R^*\right)_{\min} := \min_{(i,j)\in S_R} |\left(\Sigma_R^*\right)_{ij}|$ scales as $\left(\Sigma_R^*\right)_{\min} = \Omega\left(\sqrt{\frac{\log p}{n}}\right)$ and the sample size $n$ is lower bounded as*

$$n = \Omega\left(d^2 \log p\right), \tag{17}$$

*then with probability greater than $1 - 1/p^c \to 1$ (for some $c > 0$), we have:*

*a) The estimates $\widehat{J}_M$ and $\widehat{\Sigma}_R$ satisfy $\ell_\infty$ bounds*

$$\|\widehat{J}_M - J_M^*\|_\infty = O\left(\sqrt{\frac{\log p}{n}}\right), \tag{18}$$

$$\|\widehat{\Sigma}_R - \Sigma_R^*\|_\infty = O\left(\sqrt{\frac{\log p}{n}}\right). \tag{19}$$

*b) The estimate $\widehat{\Sigma}_R$ is sparsistent and sign consistent with $\Sigma_R^*$.*

*c) If in addition,* $\left(J_M^*\right)_{\min} := \min_{(i,j)\in E_M} \left|\left(J_M^*\right)_{ij}\right|$ *scales as* $\left(J_M^*\right)_{\min} = \Omega\left(\sqrt{\frac{\log p}{n}}\right)$*, then the estimate* $\widehat{J}_M$ *is sparsistent and sign consistent with* $J_M^*$*.*

**Remark 1 (Non-asymptotic sample complexity bounds):** In the above theorem, we establish that the number of samples is required to scale as $n = \Omega(d^2 \log p)$. In fact, the result is non-asymptotic which is provided in the extended version on arXiv.

**Remark 2 (Comparison with sparse graphical model selection):** The high dimensional covariance estimation problem which is investigated in (Ravikumar et al., 2011) involves similar mutual incoherence conditions and gives similar consistency results. Regarding the final result, sample complexity and convergence rate of estimated models are exactly the same as results in (Ravikumar et al., 2011) with only some minor differences in coefficients. But regarding the incoherence conditions, since their program is a special case of ours, the required conditions are less restrictive in their case. It is natural that we need some more incoherence conditions in order to be able to recover both the Markov and residual models.

## 5. Experiments

In this section we provide experimental results for the proposed algorithm. We term our proposed optimization program as $\ell_1 + \ell_\infty$ method and compare it with the well-known $\ell_1$ method which is a special case of the proposed algorithm when $\lambda = \infty$. The optimization programs are implemented by YALMIP (Lofberg, 2004) and SDPT3 (Toh et al., 1999) packages for MATLAB. We also compare the performance of applying belief propagation to exact models.

**Synthetic Data**: We build a Markov + residual synthetic model in the following way. The underlying graph for the Markov part is an $8 \times 8$ 2-D grid structure (4-nearest neighbor grid). We choose the off-diagonal nonzero entries in $J_M^*$ (corresponding to the grid edges) randomly from set $\{-0.5, 0.5\}$. Then we ensure that $J_M^*$ is positive definite by adding some uniform diagonal weighting. We choose 0.2 fraction of Markov edges randomly to introduce residual edges. The value of these nonzero entries in $\Sigma_R^*$ are chosen from $\{-0.2, 0.2\}$ such that the sign of residual entry is opposite of the sign of overlapping Markov entry (assumption (A.3)). We also generate a random mean in the interval $[0, 1]$ for each node.

We apply $\ell_1 + \ell_\infty$ and $\ell_1$ methods to a random realization of the above described model $\Sigma^* = {J_M^*}^{-1} + \Sigma_R^*$. The edit distance between estimated and exact Markov model $\widehat{J}_M$ and $J_M^*$ is plotted in figure 2.a for different number of samples. First observation is that by increasing the number of samples, the edit distance decreases which is consistent with theoretical results. We also see that the behaviour of $\ell_1 + \ell_\infty$ method is very close to $\ell_1$ method which suggests that sparsity pattern of $J_M^*$ can be estimated efficiently under either methods. The edit distance between $\widehat{\Sigma}_R$ and $\Sigma_R^*$ is plotted in figure 2.b. We see again decreasing trend for $\ell_1 + \ell_\infty$ method here with increasing number of samples. But since there is not any off-diagonal $\ell_\infty$ constraints in $\ell_1$ method, it can not recover the residual matrix $\Sigma_R^*$. Finally the $\ell_\infty$-elementwise norm of error between estimated precision matrix $\widehat{J}$ and the exact precision matrix $J^*$ is sketched for both methods in figure 2.c. We observe the advantage of proposed $\ell_1 + \ell_\infty$ method in estimating the overall model precision matrix $J^* = {\Sigma^*}^{-1}$.

Next the results of running Loopy Belief Propagation (LBP) on the models are presented. We compare the result of applying LBP to the same $J^*$ and $J_M^*$ models generated in the learning discussion above. The log of average mean and variance errors over all nodes are sketched in figure 3 throughout the iterations. We observe that LBP does not converge for $J^*$ model. It is shown by Malioutov et al. (2006) that if a model is walk-summable then the mean estimates under LBP converge and are accurate. The spectral norms of the partial correlation matrices are $\|\overline{R}_M\| = 0.9443$ and $\|\overline{R}\| = 6.9191$ for $J_M^*$ and $J^*$ models respectively. Thus, the matrix $J^*$ is not walk-summable and therefore its convergence under LBP is not guaranteed and this is seen in figure 3. On the other hand, LBP is accurate for $J_M^*$ matrix. Thus, our method learns models which are better suited for inference under loopy belief propagation.

## 6. Conclusion

In this paper, we provided an in-depth study of convex optimization methods and guarantees for high-dimensional covariance decomposition into sparse Markov and independence domains. We provide consistency guarantees for estimation in both the Markov and the independence domains, and establish efficient sample complexity results for our method. These findings open up many future directions to explore. One important aspect is to relax the sparsity constraints imposed in the two domains, and to develop new methods to enable decomposition of such models. Other considerations include extension to discrete models and other models for the residual covariance matrix (e.g. low rank matrices). Such findings will push the envelope of efficient models for high-dimensional estimation. It is worth mentioning while in many scenarios it is important to incorporate latent variables, in our framework it is challenging to incorporate both latent variables as well as marginal independencies, and provide learning guarantees, and we defer it to future work.

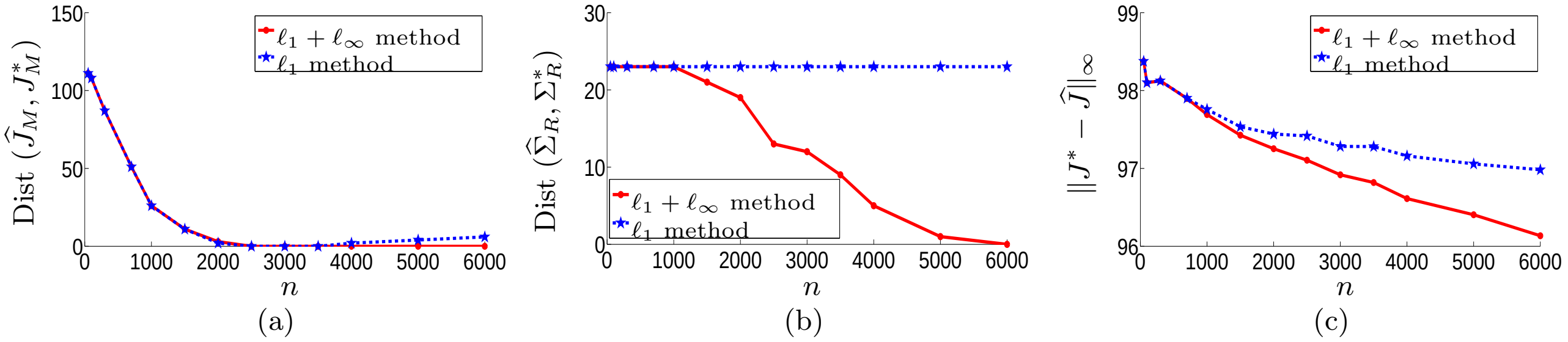

*Figure 2.* (a) Edit distance between estimated Markov model $\widehat{J}_M$ and exact Markov model $J^*_M$. (b) Edit distance between estimated residual model $\widehat{\Sigma}_R$ and exact residual model $\Sigma^*_R$. (c) precision matrix estimation error $\|J^* - \widehat{J}\|_\infty$ , where $\widehat{J} = \widehat{J}_M$ for $\ell_1$ method and $\widehat{J} = \left(\widehat{J}_M^{-1} + \widehat{\Sigma}_R\right)^{-1}$ for $\ell_1 + \ell_\infty$ method.

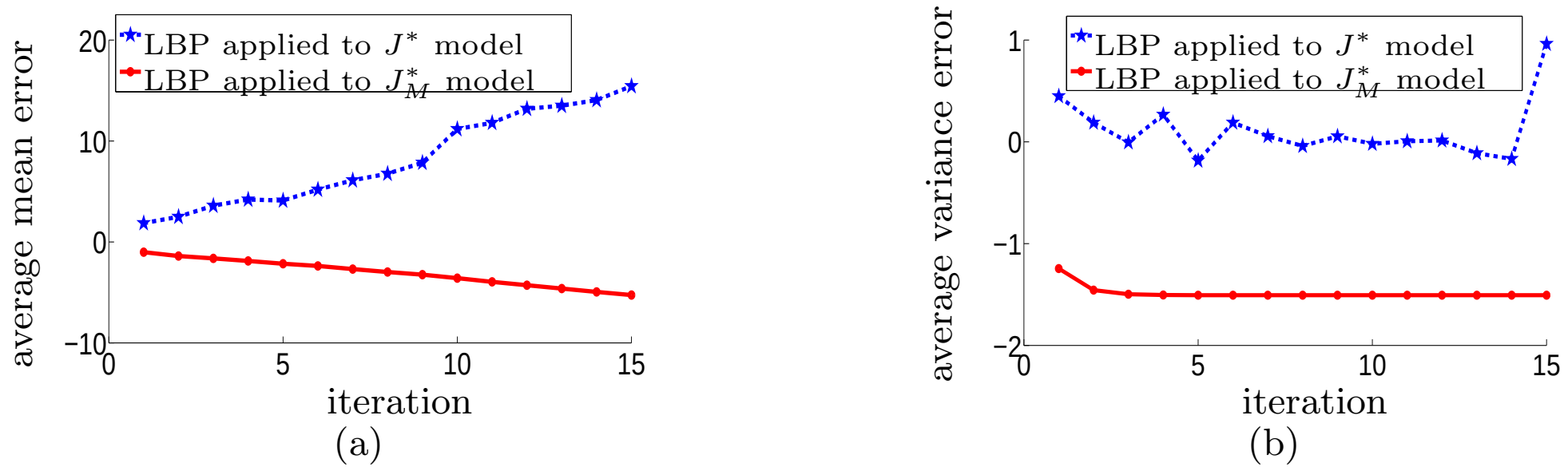

*Figure 3.* Performance under loopy belief propagation for the overall model ($J^*$) and the Markov component ($J^*_M$).

## References

Anderson, T.W. *An Introduction to Multivariate Statistical Analysis*. John Wiley & Sons, Inc., New York, NY, 1984.

Boyd, S.P. and Vandenberghe, L. *Convex Optimization*. Cambridge University Press, 2004.

Brown, L.D. Fundamentals of statistical exponential families: with applications in statistical decision theory. *Lecture Notes-Monograph Series, Institute of Mathematical Statistics*, 9, 1986.

Choi, M.J., Chandrasekaran, V., and Willsky, A.S. Gaussian multiresolution models: Exploiting sparse Markov and covariance structure. *Signal Processing, IEEE Transactions on*, 58(3):1012–1024, 2010.

Cover, T. and Thomas, J. *Elements of Information Theory*. John Wiley & Sons, Inc., 2006.

Kauermann, G. On a dualization of graphical gaussian models. *Scandinavian journal of statistics*, pp. 105–116, 1996.

Lauritzen, S.L. *Graphical models: Clarendon Press*. Clarendon Press, 1996.

Lofberg, J. Yalmip: A toolbox for modeling and optimization in matlab. In *IEEE international symposium on Computer Aided Control Systems Design (CACSD)*, September 2004. Available from http://users.isy.liu.se/johanl/yalmip/.

Malioutov, D.M., Johnson, J.K., and Willsky, A.S. Walk-Sums and Belief Propagation in Gaussian Graphical Models. *J. of Machine Learning Research*, 7:2031–2064, 2006.

Ravikumar, P., Wainwright, M.J., Raskutti, G., and Yu, B. High-dimensional covariance estimation by minimizing $\ell_1$-penalized log-determinant divergence. *Electronic Journal of Statistics*, (4):935–980, 2011.

Toh, K. C., Todd, M.J., and Tutuncu, R. H. Sdpt3 - a matlab software package for semidefinite programming. *Optimization Methods and Software*, 11:545–581, 1999.

Wainwright, M.J. and Jordan, M.I. Graphical Models, Exponential Families, and Variational Inference. *Foundations and Trends in Machine Learning*, 1(1-2):1–305, 2008.